\documentclass[a4paper,twoside]{article}

\usepackage{epsfig}
\usepackage{subfigure}
\usepackage{calc}
\usepackage{amssymb}
\usepackage{amstext}
\usepackage{amsmath}
\usepackage{amsthm}
\usepackage{multicol}
\usepackage{pslatex}
\usepackage{apalike}
\usepackage{subfigure}
\usepackage{balance}
\usepackage{SCITEPRESS}     
\usepackage{url}
\begin{document}

\title{Automatic 3D Point Set Reconstruction from Stereo Laparoscopic Images using Deep Neural Networks}

\author{\authorname{B\'alint Antal\sup{1}}
\affiliation{\sup{1}Faculty of Informatics , University of Debrecen, Debrecen, Hungary}
\email{antal.balint@inf.unideb.hu}
}


\abstract{In this paper, an automatic approach to predict 3D coordinates from stereo laparoscopic images is presented. The approach maps a vector of pixel intensities to 3D coordinates through training a six layer deep neural network. The architectural aspects of the approach is presented and in detail and the method is evaluated on a publicly available dataset with promising results.}

\onecolumn \maketitle \normalsize \vfill

\section{\uppercase{Introduction}}
\label{sec:introduction}

\begin{figure*}[htb]
\centering
\subfigure[Left image]{
\label{fig:no}
\includegraphics[keepaspectratio,width=0.4\linewidth]{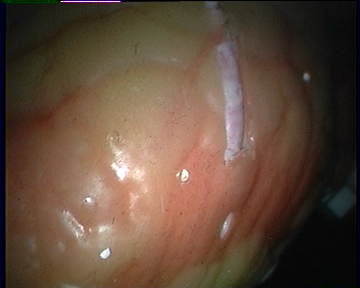}
}
\quad
\subfigure[Right image]{
\label{fig:bs}
\includegraphics[keepaspectratio,width=0.4\linewidth]{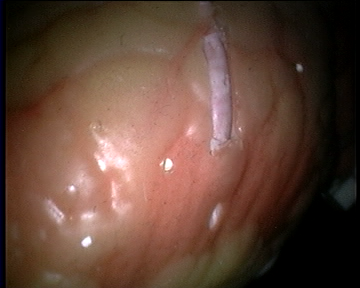}
}\caption{An example stereo laparoscopy image pair \cite{pratt2010} \cite{stoyanov2010}.}
\label{fig:left_right}
\end{figure*}

\noindent Minimally invasive surgery (MIS) became a wide-spread technique to have surgical access to the abdomen of patients without casing major damages in the skin or tissues. Since MIS supporting techniques like laparascopy or endscopy provide a restricted access to the surgeon, computer-aided visualisation systems are developed. One of the major research areas in the 3d reconstruction of stereo endoscope images. See Figure \ref{fig:left_right} for an example stereo cardiac laparoscopy image pair. 

\begin{figure*}[htb]
\centering
\subfigure[Disparsity map]{
\label{fig:disp}
\includegraphics[keepaspectratio,width=0.4\linewidth]{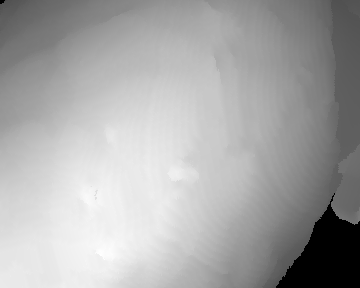}
}
\quad
\subfigure[Distance map]{
\label{fig:dm}
\includegraphics[keepaspectratio,width=0.4\linewidth]{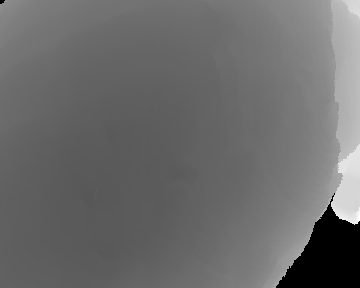}
}
\\
\subfigure[Reconstructed 3D point cloud]{
\label{fig:pc}
\includegraphics[keepaspectratio,width=0.4\linewidth]{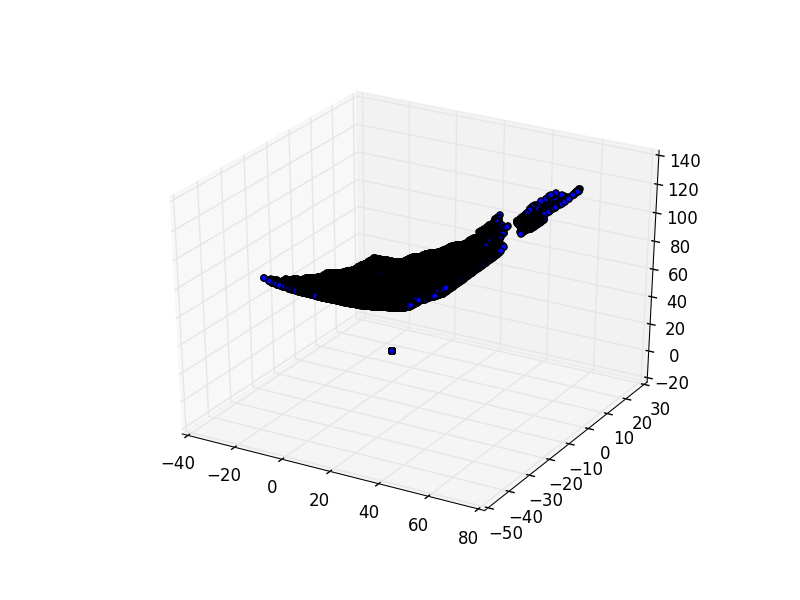}
}
\caption{Disparsity map, distance map and 3D point cloud extracted from the images shown in Figure \ref{fig:left_right}.}
\label{fig:gt}
\end{figure*}

The images are acquired from two distinct viewpoints, assisting surgeons to have a sense of depth during surgery. The usual 3d reconstruction approach consists of several steps \cite{optical}, involving the establishment of stereo correspondence between the pixels of the two viewpoints, which is a computationally expensive approach and also requires prior knowledge regarding the endoscope used in the procedure, limiting its reusability. Figure \ref{fig:gt} shows an example stereo image pair from the \cite{pratt2010} \cite{stoyanov2010}, alongside the reconstructed disparsity map, distance map and 3d point cloud.
In this paper, an automatic approach for the 3d reconstruction of stereo endoscopic images will be presented. The approach is based on deep neural networks (DNN) \cite{deep_learning} and aims to predict 3d coordinates without the costly procedure of stereo correspondence. We have evaluated our approach on a publicly available database where it performed well compared to a stereo correspondence approach.   

\begin{figure*}
\centering
\includegraphics[keepaspectratio,width=0.5\linewidth]{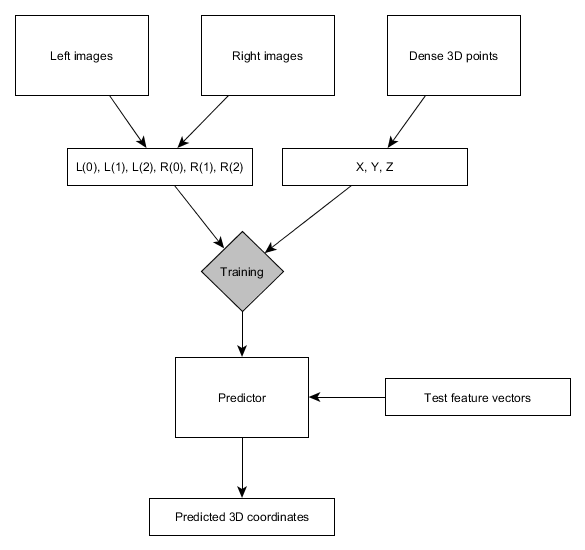}
\caption{Flow chart of the proposed approach.}
\label{fig:flowchart}
\end{figure*}

The proposed approach only takes the pixel intensity values for the left and right images, and learns their 3D depth map during training. 
 Figure \ref{fig:flowchart} shows the flow chart of the proposed approach. 

The rest of the paper is organized as follows: section \ref{sec:approach} describes the proposed approach in details. We provide our experimental methodology in section \ref{sec:methodology}. Section \ref{sec:results} contains the results of our experiments and finally, we draw conclusion in section \ref{sec:conclusion}.
 
\section{\uppercase{3D Reconstruction of stereo endoscopic images using Deep Neural Networks}}
\label{sec:approach}

In this section, the proposed approach is described in details. Section \ref{sec:dnn} presents an overview on DNNs. Section \ref{sec:arch} proposes the architecture of our approach, while we describe the optimization aspects of the proposed method in section \ref{sec:opt}. 

\subsection{Deep Neural Networks}
\label{sec:dnn}

Deep Neural Networks (DNNs) are biologically inspired machine learning techniques which does not rely on engineering domain-specific features for each separate problem but involves a mapping of the input data (e.g. images) to a target vector using a sequence of non-linear transformations \cite{deep_learning}. In particular, DNNs consists of several layers of high-level representations on the input data. Each layer consists of several nodes, which contains weights, biases and activations in the following form:
\begin{equation}
	out\left(x\right) = \left(f\left(W \cdot x + b\right)\right), 
\end{equation}
where $x$ is an input vector, $f$ is a non-linear activation function. $W$ is a weight matrix of shape $N \times M$, $N$ and $M$ are the  output and input dimensions of the preceding and succeeding layers, respectively, $b$ is a bias vector.  
Each DNN has an input layer, an output layer and several hidden layers of this form. DNN has proven to be very successful in a wide variety of machine learning related tasks. 

\subsection{Architecture of the proposed DNN}
\label{sec:arch}

\begin{figure*}
\centering
\includegraphics[keepaspectratio,width=0.8\linewidth]{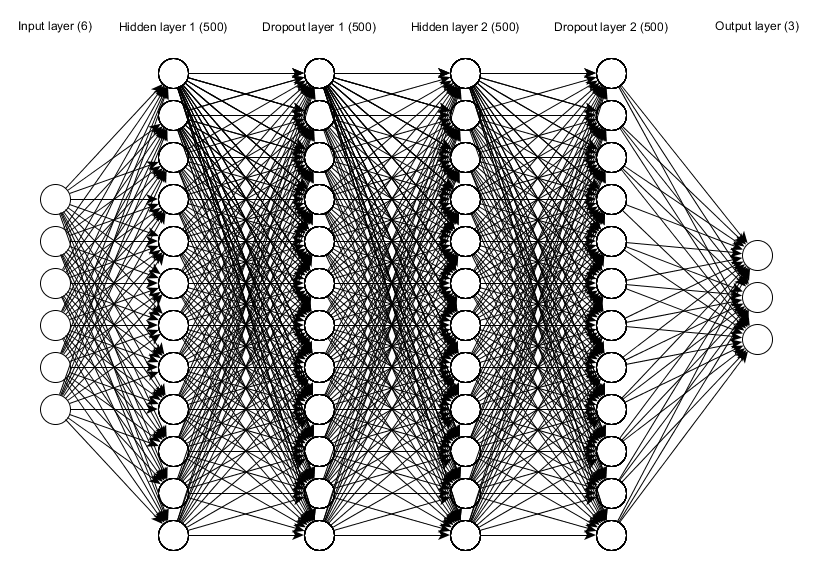}
\caption{Architecture of the proposed DNN.}
\label{fig:architecture}
\end{figure*}

The proposed DNN consists of six layers (see Figure \ref{fig:architecture}). 
The input layer takes a six element vector of the following form: $
	x_{i,j} = \left(L_{i,j}\left(0\right),\, L_{i,j}\left(1\right),\, L_{i,j}\left(2\right),\,  R_{i,j}\left(0\right),\, R_{i,j}\left(1\right),\, R_{i,j}\left(2\right)\right),
$ where $L_{i,j}\left(C\right)$ and $R_{i,j}\left(C\right)$ are the pixel intensites at the $i,j$ position in channel $C = \{0,1,2\}$ of the left and right images, respectively. 
The output layer produces three outputs, namely the $X$, $Y$, $Z$ coordinates of the input points. Two fully connected dense layers serves as hidden layers in our architecture with 500 neurons each. A high-dimensional upscaling of the input features like this are effective in learning complex mappings. To avoid overfitting, a dropout layer after each dense layers were also included, which introduces random noise into the outputs of each layer \cite{dropout}.
We have used Rectified Linear Units as activiations, \cite{relu} \cite{glorot2011} which is of the form:
\begin{equation}
	f\left(x\right) = max\left(0,\,x\right).
\end{equation}

\subsection{Optimization}
\label{sec:opt}

We have used an adaptive per-dimension learning rate version of the stochastic gradient descent (SGD) approach called Adadelta, which is less sensitive to hyperparameter settings than SGD \cite{adadelta}. Each $W$ weight matrix and $b$ vector were initialized using the normalized initialization \cite{glorot2010}:
\begin{equation}
	U\left|-\dfrac{\sqrt{6}}{\sqrt{n_{j}+n_{j+1}}},\,\dfrac{\sqrt{6}}{\sqrt{n_{j}+n_{j+1}}}\right|,
\end{equation}
where U is the uniform distribution, $n_{j}$ and $$n_{j+1}$$ are the sizes of the previous and next layers, respectively. To avoid overfitting, we have also applied Tikhonov regularization for each weight matrix \cite{ridge_regression}.
As an energy function, we have used mean squared error:
\begin{equation}
	MSE = \dfrac{\sum_{N}\left(\left(x-x'\right)^2 + \left(y-y'\right)^2\left(z-z'\right)^2\right)}{N},
\end{equation}
where $x$, $y$ and $z$ are the ground truth coordinates and $x'$, $y'$ and $z'$ are the DNNs predictions, and $N$ is the number of training vectors, respectively.

\section{\uppercase{Methodology}}
\label{sec:methodology}

We have used a laparascpic cardiac dataset to evaluate our approach \cite{pratt2010} \cite{stoyanov2010}. The dataset consists a pair of videos showing heart movement. The video consist of 2427 frames, each of them having a spatial resolution of $360\times288$ and in a standard RGB format. The ground truth is a depth map containing $X$, $Y$, $Z$ coordinates for each point. 
We have used a training set of 20 images and tested our approach on the remaining 2407 frames. For training, we have allowed 20 epochs to establish the optimal parameters for the DNN. After training, we have applied pixel-wise classification of the images of the test set.We have calculated the root mean squared errors for each image:
\begin{equation}
	RMSE = \dfrac{\sum_{N}\left(\sqrt{\left(x-x'\right)^2 + \left(y-y'\right)^2\left(z-z'\right)^2}\right)}{N},
\end{equation}
 
We have used Theano \cite{theano2010} \cite{theano2012} and Keras \cite{keras} for implementation.  

\section{\uppercase{Results and discussion}}
\label{sec:results}

\begin{figure*}
\centering
\includegraphics[keepaspectratio,width=0.8\linewidth]{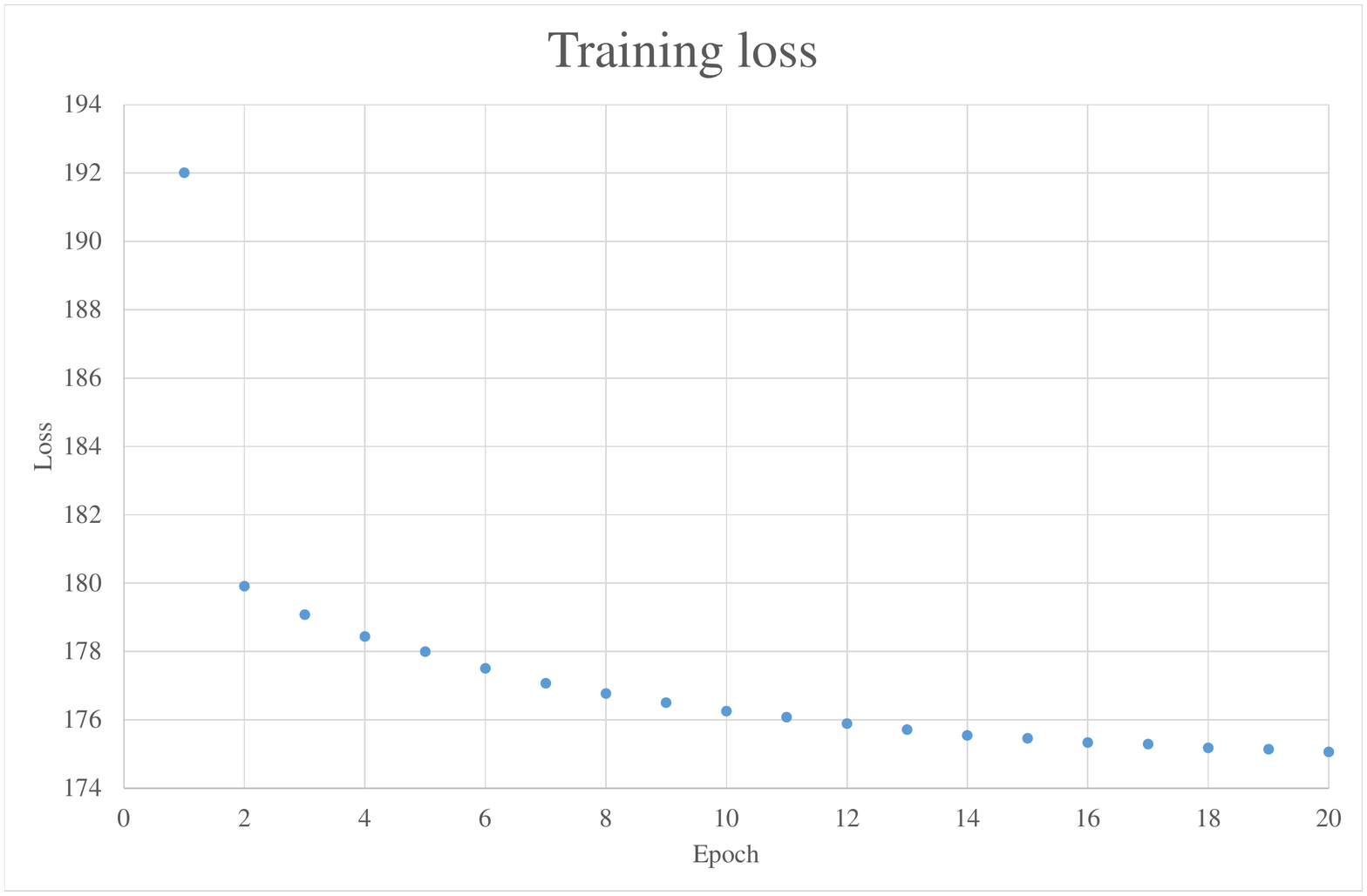}
\caption{Training losses per epoch.}
\label{fig:training_loss}

\end{figure*}

First, to show the validity of the DNN architecture and the training hyperparameters, we have measured the training loss at each epoch. As it can be seen in Figure \ref{fig:training_loss}, the training loss decreased in each epoch showing but the curve started to flatten at later epochs. 

\begin{figure}
\centering

\includegraphics[keepaspectratio,width=0.8\linewidth]{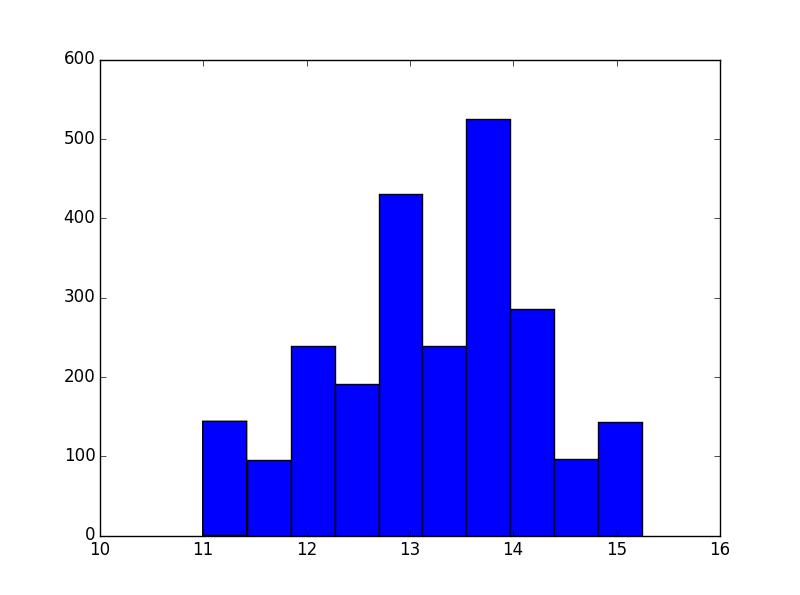}
\caption{Histogram of root mean squared error losses on the test data.}
\label{fig:test_loss}
\end{figure}

We have also calculated the loss on the test instances. In average, the  RMSE per image was 13.18. As it can be seen from Figure \ref{fig:test_loss}, there was quite a big fluctuation in losses (around 11-15). However, the ground truth was incomplete for some of the images, which might have affected the ability to properly evaluate each individual instances.

\begin{figure*}[htb]
\centering

\subfigure[Point cloud from ground truth]{
\label{fig:pcgt}
\includegraphics[keepaspectratio,width=0.4\linewidth]{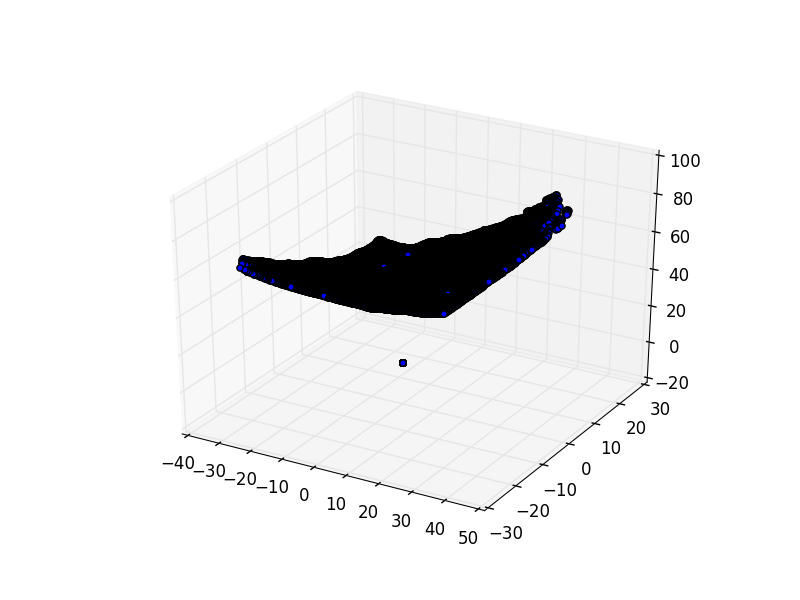}
}
\quad
\subfigure[Point cloud predicted by the proposed method]{
\label{fig:pcpred}
\includegraphics[keepaspectratio,width=0.4\linewidth]{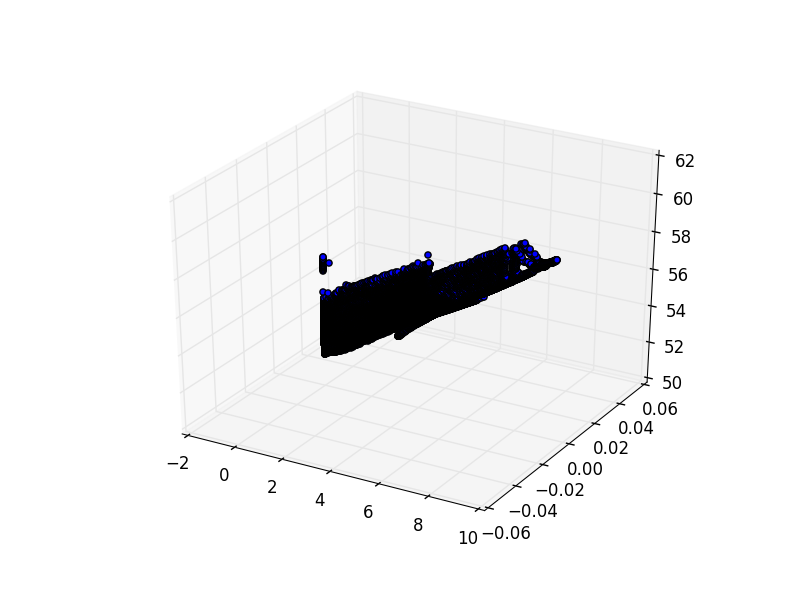}
}

\caption{Comparison of point clouds extracted from the ground truth and by our approach.}
\label{fig:comparison}
\end{figure*}

However, in some extreme cases (see Figure \ref{fig:comparison}), the reconstruction by the proposed approach was just partially successful. To correct issues like this, an approach which also incorporates on contextual information could be used in the future.

\section{\uppercase{Conclusions}}
\label{sec:conclusion}

Minimally invasive surgical techniques are very important in clinical settings, however, they require computational support to allow surgeons to effectively use these techniques in practice. In this paper, an approach based on deep neural networks has been introduced, which is unlike the state-of-the-art approaches, only relies on the input pixels  of the stereo image pair. The approach has been evaluated on a publicly available dataset and compared well to the results obtained by a state-of-the-art technique.

\section*{\uppercase{Acknowledgments}}

This work was supported in part by the project VKSZ 14-1-2015-0072, SCOPIA: Development of diagnostic tools based on endoscope technology supported by the European Union, co-financed by the European Social Fund.

\vfill
\balance
\bibliographystyle{apalike}
{\small
\bibliography{refs}}

\vfill
\end{document}